\documentclass[11pt,a4paper]{article}
\usepackage[hyperref]{emnlp2020}
\usepackage{times}
\usepackage{latexsym}

\usepackage{amsfonts}
\usepackage{amsmath}
\usepackage{multirow}
\usepackage{graphicx}  
\usepackage{booktabs} 
\usepackage{tabularx}

\usepackage{microtype}
\usepackage{subfigure}

\aclfinalcopy 
\def\aclpaperid{***} 

\title{Exploring and Evaluating Attributes, Values, and Structures for \\ Entity Alignment}

\author{Zhiyuan Liu$^{12}$ \quad Yixin Cao$^2$\footnotemark \quad Liangming Pan$^{12}$\\
\textbf{Juanzi Li$^3$ \quad Zhiyuan Liu$^3$ \quad Tat-Seng Chua$^2$}\\
$^1$NUS Graduate School for Integrative Sciences and Engineering\\
$^2$School of Computing, National University of Singapore, Singapore\\
$^3$Department of CST, Tsinghua University, Beijing, China\\
{\tt \{acharkq, caoyixin2011\}@gmail.com, e0272310@u.nus.edu} \\
{\tt \{lijuanzi, liuzy\}@tsinghua.edu.cn, dcscts@nus.edu.sg}
}

\date{}

\begin{document}
\maketitle
\begin{abstract}
    Entity alignment (EA) aims at building a unified Knowledge Graph (KG) of rich content by linking the equivalent entities from various KGs. GNN-based EA methods present promising performance by modeling the KG structure defined by relation triples. However, attribute triples can also provide crucial alignment signal but have not been well explored yet. In this paper, we propose to utilize an attributed value encoder and partition the KG into subgraphs to model the various types of attribute triples efficiently. Besides, the performances of current EA methods are overestimated because of the name-bias of existing EA datasets. To make an objective evaluation, we propose a hard experimental setting where we select equivalent entity pairs with very different names as the test set. Under both the regular and hard settings, our method achieves significant improvements ($5.10\%$ on average Hits@1 in DBP15k) over 12 baselines in cross-lingual and monolingual datasets. Ablation studies on different subgraphs and a case study about attribute types further demonstrate the effectiveness of our method. Source code and data can be found at \url{https://github.com/thunlp/explore-and-evaluate}.
\end{abstract}

\section{Introduction}
{\renewcommand{\thefootnote}{*}
\footnotetext{Corresponding author.}}

The prosperity of data mining has spawned Knowledge Graphs (KGs) in many domains that are often complementary to each other. Entity Alignment (EA) provides an effective way to integrate the complementary knowledge in these KGs into a unified KG by linking equivalent entities, thus benefiting knowledge-driven applications such as Question Answering~\cite{yang2017efficiently,yang2018hotpotqa}, Recommendation~\cite{cao2019unifying} and Information Extraction~\cite{kumar2017survey, cao2018neural}. 
However, EA is a non-trivial task that it could be formulated as a quadratic assignment problem~\cite{yan2016short}, which is NP-complete~\cite{garey1990computers}.

\begin{figure*}[!t]
    \centering
    \subfigure[EA by the structure assumption.]{
        \label{fig:struc}\includegraphics[width=0.32\textwidth]{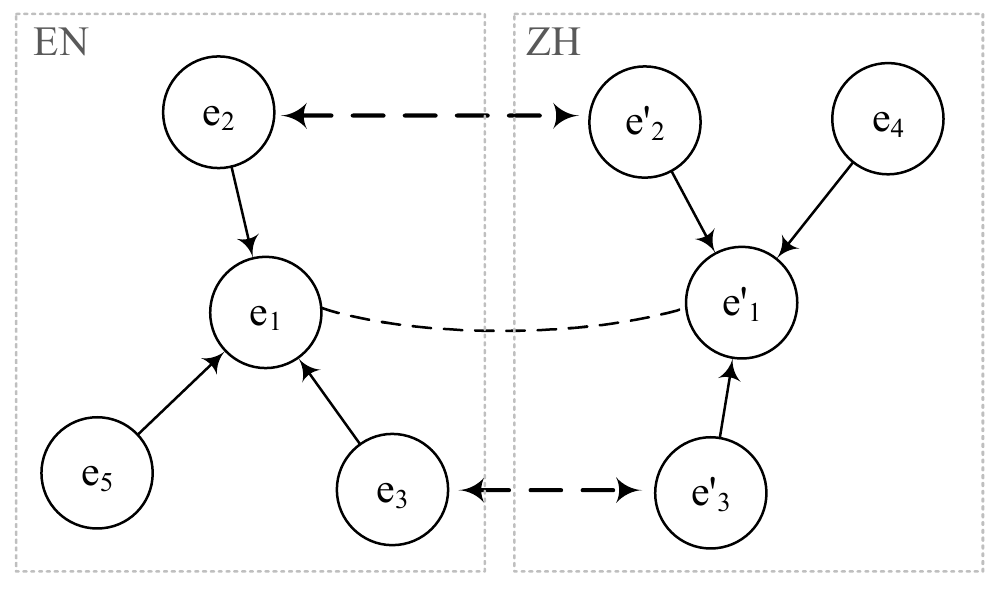}}
    \subfigure[EA by the attribute assumption.]{
        \label{fig:value}\includegraphics[width=0.32\textwidth]{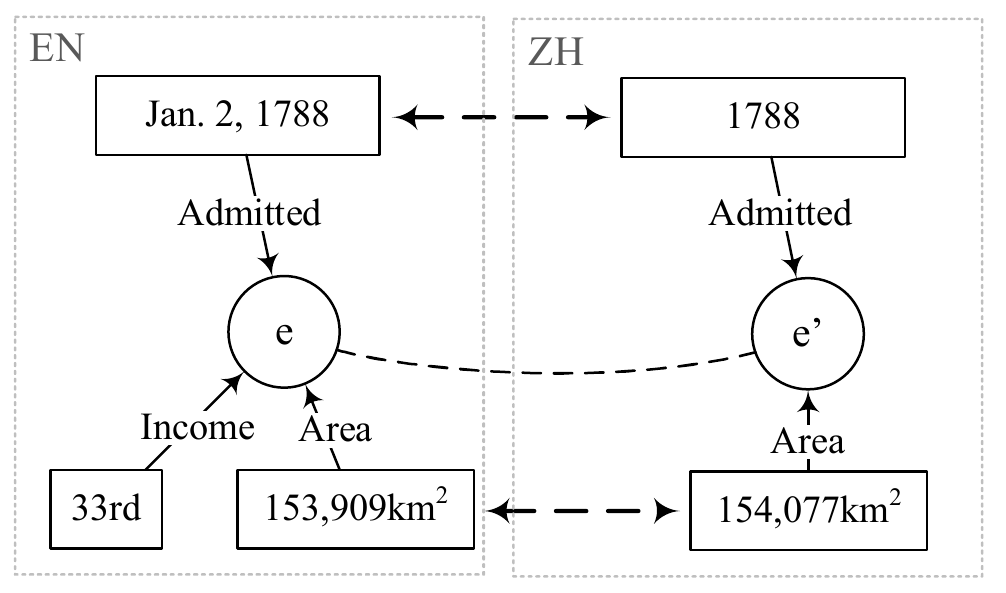}}
    \subfigure[EA with attribute importance.]{
        \label{fig:attr}\includegraphics[width=0.32\textwidth]{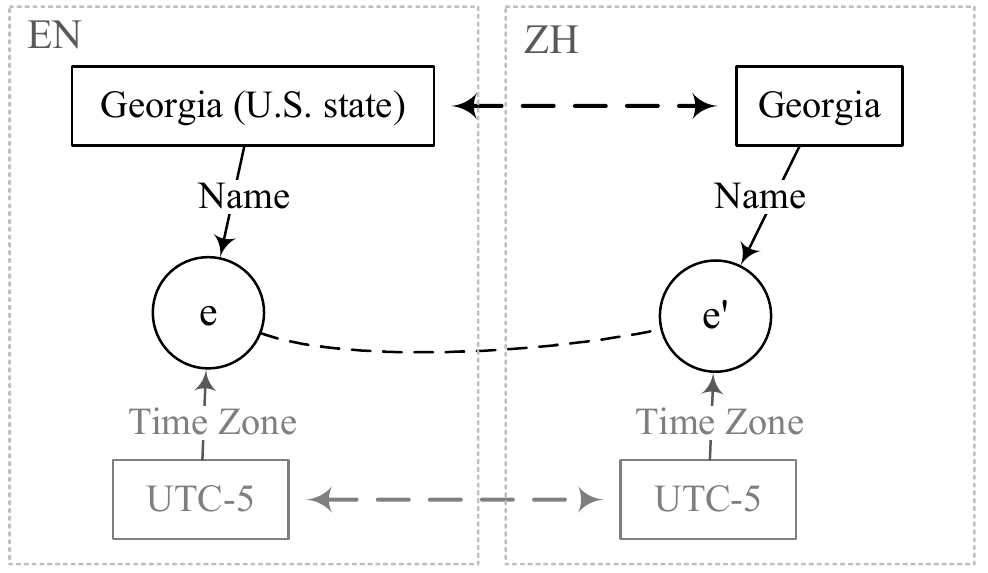}}
    \caption{Examples for EA using different assumptions and identifying the different importance of attributes. In Figure~\ref{fig:struc}, we align $e_1$ and $e'_1$ for the equivalent entity pairs $(e_2, e_2')$ and $(e_3, e_3')$ in their neighbors. In Figure~\ref{fig:value},\ref{fig:attr}, we align $e$ and $e'$ for their similar attributes and values; $e$ refers to the entity ``Georgia (U.S. state)'' from English Wiki and $e'$ is the Chinese equivalent. In Figure~\ref{fig:attr}, attribute \textit{Time Zone} and its value is assigned less attention weight for being less discriminative for alignment. Chinese texts are translated. Dashed curves link the target equivalent entity pairs. Dashed bothway arrows indicate alignment signals.}
    \label{fig:example}
\end{figure*}

A KG comprises a set of triples, with each triple consisting of a {\it subject}, {\it predicate}, and {\it object}. There are two types of triples: (1) \textit{relation triples}, in which both the subject and object are entities, and the predicate is often called \textit{relation} (see Figure~\ref{fig:struc}); and (2) \textit{attribute triples}, in which the subject is an entity and the object is a \textit{value}, which is either a number or literal string (see Figure~\ref{fig:attr}), and the predicate is often called \textit{attribute}.

Most of the previous EA models~\cite{sun2017cross, wang2018cross, wu2019relation} rely on the structure assumption that, the adjacencies of two equivalent entities in KGs usually contain equivalent entities~\cite{wang2018cross} (see Figure~\ref{fig:struc}). These models mainly focus on modeling KG structure defined by the relation triples. However, we argue that attribute triples can also provide important clues for judging whether two entities are the same, based on the attribute assumption that:
{\it equivalent entities often share similar attributes and values in KGs.}
For example, in Figure~\ref{fig:value}, the equivalent entities $e$ and $e'$ share the attribute \textit{Area} with similar values of $153,909$ and $154,077$. 
Therefore, we aim to improve EA using attribute triples. We have identified the challenges of attribute incorporation and dataset bias.

\textbf{Attribute Incorporation Challenge.} 
Modeling attribute triples together with relation triples is a more effective strategy than modeling attribute triples alone. In this way, the alignment signal from attribute triples can be propagated to an entity's neighbors via relation triples. Recently, some pioneer EA works~\cite{Zhang2019MultiviewKG,trisedya2019entity} have incorporated both attribute and relation triples. However, they learn relation and attribute triples in separate networks. In this case, the alignment signal from an entity's discriminative attributes and values will be reserved to the entity itself and will not help align its neighbors.
In addition, it is crucial to identify the different importance of attributes in discriminating whether two entities are equivalent. For example, the attribute \textit{Time Zone} should be assigned less importance than \textit{Name} since many cities can share the same \textit{Time Zone} (Figure~\ref{fig:attr}). Previous works fail to consider the different importance of attributes.

\textbf{Dataset Bias Challenge.} The performance of EA is overestimated because the existing EA datasets are biased to the attribute \textit{Name}: $60\%-80\%$ of the released seed set of equivalent entities in DBP15k can be aligned via name matching. The reason is that the equivalent entities are collected using inter language links, which are labeled by a strategy that heavily relies on the translation of entity names\footnote{https://en.wikipedia.org/wiki/Help:Interlanguage\_links}.
In this way, the datasets contain many ``easy'' equivalent entities that have similar names.
However, in the practical application of EA, the ``easy'' equivalent entities are often aligned already, and the challenge is to align the ``hard'' ones that have very different names. 
This discrepancy between datasets and practical situation causes overestimated EA performance.

To address the first challenge, we propose \textbf{Attr}ibuted \textbf{G}raph \textbf{N}eural \textbf{N}etwork (AttrGNN) to learn attribute triples and relation triples in a unified network, and learn importance of each attributes and values dynamically. Specifically, we propose an \textit{attributed value encoder} to select and aggregate alignment signal from informative attributes and values. We further employ the mean aggregator~\cite{hamilton2017inductive} to propagate this signal to entity's neighbors. In addition, as different types of attributes have different similarity measurements, we partition the KG into four subgraphs by grouping attributes, \textit{i.e.}, attribute \textit{Name}, literal attribute, digital attribute, and structural knowledge. We apply separate channels to learn their representations. We present two methods to ensemble the outputs from all channels.

To alleviate the name-bias of EA datasets (second challenge), we propose a hard experimental setting. Specifically, we construct harder test sets from existing datasets by selecting equivalent entities that have the least similarity in their names. We further evaluate the models on these harder test sets to offer a more objective evaluation of EA models' performance. Under both the hard and regular settings, AttrGNN achieves the best result with significant performance improvement ($5.10\%$ Hits@1 on average in DBP15k) over 12 baselines on both the cross-lingual and monolingual datasets.

\section{Related Work}
Recent entity alignment methods can be classified into embedding-based methods and Graph Neural Network-based (GNN-based) methods.

\begin{figure*}[!t]
    \vspace{-0.2cm}
    \centering
    \includegraphics[width=0.98\textwidth]{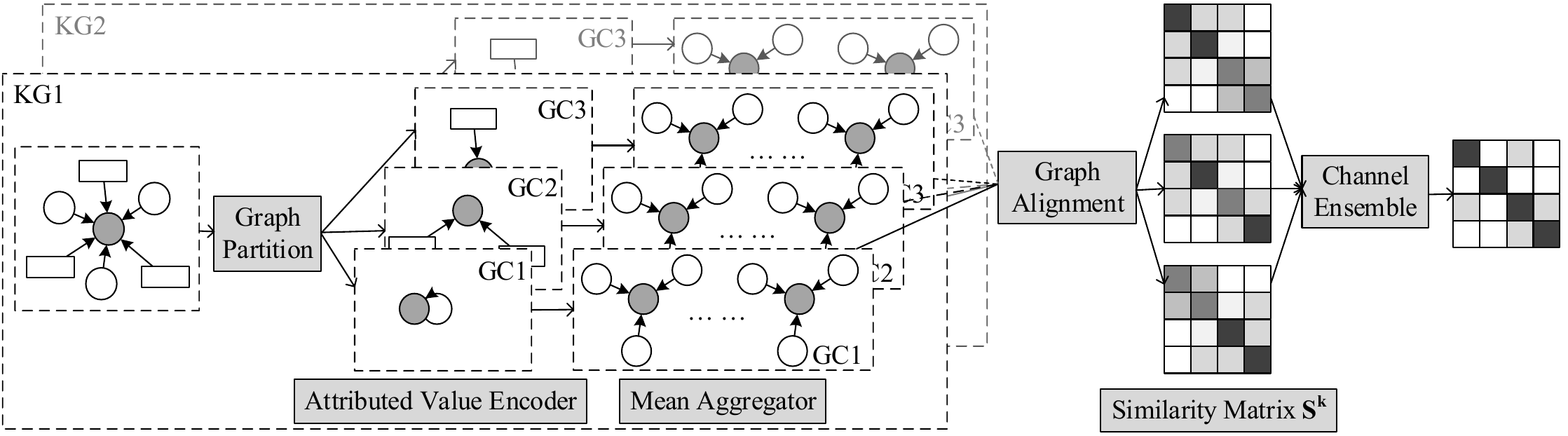}
    \caption{The framework of AttrGNN. Three GNN channels (GCs) are shown as an example. We do not use any attributes in GC1 to focus on the learning of structural knowledge (node degree distribution). $\mathbf{S}^k$ is the output similarity matrix of GC$k$. $\mathbf{S}^k_{e, e'}$ is the similarity between $e\in$ KG1 and $e'\in$ KG2 measured by GC$k$. For the KGs and its subgraphs, we use circles to denote entities and rectangles to denote values.}
    \label{fig:frame}
    \vspace{-0.2cm}
\end{figure*}

\subsection{Embedding-based Methods}
Recent works utilize KG embedding methods, such as TransE~\cite{bordes2013translating}, to model the relation triples and further unifies two KG embedding spaces by forcing seeds to be close~\cite{chen2016multilingual}. 
Attribute triples has been introduced in this field. JAPE~\cite{sun2017cross} computes attribute similarity to regularize the structure-based optimization. KDCoE~\cite{chen2018co} co-trains entity description and structure embeddings with a shared iteratively enlarged seed set. AttrE~\cite{trisedya2019entity} and MultiKE~\cite{Zhang2019MultiviewKG} encode values as extra entity embeddings. However, the diversity of attributes and uninformative values limit the performance of the above methods.

\subsection{GNN-based Methods}

Following Graph Convolutional Networks~\cite{kipf2016semi}, many GNN-based models are proposed because of GNN's strong ability to model graph structure. These methods present promising results on EA because GNN can propagate the alignment signal to the entity's distant neighbors. Previous GNN-based methods focus on extending GNN's ability to model relation types~\cite{wu2019relation, wu2019jointly, li2019semi}, aligning entities via matching subgraphs~\cite{Xu2019CrosslingualKG, wu2020neighborhood}, and reducing the heterogeneity between KGs~\cite{cao2019multi}. With the exception of~\citet{wang2018cross} that have incorporated attributes as the initial feature of entities, most of the current GNN-based methods fail to incorporate the attributes and values to further improve the performance of EA. 

In this paper, we add values as nodes into graph and use an attributed value encoder to conduct attribute-aware value aggregation. 


\section{Methodology}
The key idea of AttrGNN is to use graph partition and attributed value encoder to deal with various types of attribute triples. In this section, we first define KG and then introduce our graph partition strategy. Further, we design different GNN channels for different subgraphs and present two methods to ensemble all channels' outputs for final evaluation.

\subsection{Model Framework}


\noindent \textbf{Knowledge Graph (KG)} is formalized as a 6-tuple directed graph $G=(E,R,A,V,T^r,T^a)$ where $E$, $R$, $A$, and $V$ refer to the set of entities, relations, attributes, and values, respectively. $T^r=\{(h,r,t) \ \vert \ h, t \in E, r \in R\}$ and $T^a=\{(e,a,v) \ \vert \ e \in E, a \in A, v \in V)\}$ is the set of relation triples and attribute triples.

\noindent\textbf{Entity Alignment} is to find a mapping between two KGs $G$ and $G'$, \textit{i.e.}, $\psi=\{(e,e') \ \vert \ e\in E, e'\in E'\}$, where $e$ and $e'$ are equivalent entities. A seed set of equivalent entities $\psi^s$ is used as training data.

\noindent\textbf{Framework.} The framework of our AttrGNN model is shown in Figure~\ref{fig:frame}, which consists of four major components: (1) \textit{Graph Partition}, which divides the input KG into subgraphs by grouping attributes and values. (2) \textit{Subgraph Encoder}, which employs multiple GNN channels to learn the subgraphs separately. Each channel is a stack of $L$ attributed value encoders and mean aggregators. The attributed value encoder aggregate attributes and values to generate the entity embeddings, and the mean aggregator propagates entity features to its neighbors following the graph structure. (3) \textit{Graph Alignment}, which unifies the entity vector spaces of two KGs for each channel. (4) \textit{Channel Ensemble}, which infers the entity similarity using each channel and ensemble all channels' results for final inference.

\subsection{Graph Partition}
Attributes and values have various types, \textit{e.g.}, strings $\mathbb{S}$ and numbers $\mathbb{R}$. Different attributes have different similarity measurements, for example, the similarity between digital values should be numerical differences ($153,909$ \textit{v.s.} $154,077$), while the similarity of literal values is often based on their semantic meanings. Therefore, we separately learn the similarity measurements of the KG's 4 subgraphs, defined as $G^{k}=(E,R,A^k,V^k,T^r,T^{ak})$, where $k\in\{1, 2, 3, 4\}$:

\noindent $\bullet$ $G^{1}$ includes attribute triples of \textit{Name} only, \textit{i.e.}, $A^1=\{a_{name}\}$. 

\vspace{0.1cm}

\noindent $\bullet$ $G^{2}$ includes attribute triples of literal values, \textit{i.e.}, $A^2=\{a\ \vert \ (e,a,v)\in T^a,v\in \mathbb{S}, a \neq a_{name} \}$. 

\vspace{0.1cm}

\noindent $\bullet$ $G^{3}$ includes attribute triples of digital values, \textit{i.e.}, $A^3=\{a\ \vert \ (e,a,v)\in T^a,v\in \mathbb{R}\}$;

\vspace{0.1cm}

\noindent $\bullet$ $G^{4}$ has no attribute triples, \textit{i.e.}, $A^4=\emptyset$.

\vspace{0.1cm}

These subgraphs have mutually-exclusive attribute triples but share the same relation triples.

\subsection{Subgraph Encoder}
\label{sec:channel}
We design different GNN channels (GCs) to encode the above four subgraphs: {\it Name} channel for $G^1$, {\it Literal} channel for $G^2$, {\it Digital} channel for $G^3$, and {\it Structure} channel for $G^4$. The building blocks of these channels are two types of GNN layers: the attributed value encoder and the mean aggregator. Particularly, to select alignment signal from the informative attributes and values, we first stack one attributed value encoder and then mean aggregators in the {\it Literal} and {\it Digital} channels. We stack no attributed value encoder and only mean aggregators for the {\it Structure} and {\it Name} channels because they do not use various attribute triples. We add residual connections~\cite{he2016deep} between GNN layers for the {\it Name}, {\it Literal}, and {\it Digital} channels. Following previous EA works, all channels have two GNN layers. Next, we describe attributed value encoder and mean aggregator in details.

\subsubsection{Attributed Value Encoder}
Attributed value encoder can selectively gather discriminative information from the initial feature of attributes and values to the central entity. As an example, we show how to obtain e's first layer hidden state $\mathbf{h}_e^1$. The same method applies to all the entities. We obtain the sequence of attribute features $\{\mathbf{a}_1, \cdots, \mathbf{a}_n\}$ and value features $\{\mathbf{v}_1, \cdots, \mathbf{v}_n\}$ given the attribute triples $\{(e, a_1,v_1),\cdots, (e, a_n, v_n)\}$ of $e$ as inputs. Specifically, we use BERT~\cite{devlin2019bert} to obtain the features of both literal and digital values\footnote{As shown by~\citet{andor2019giving}, BERT embedding can be used for simple numerical computation.}. 
BERT is a language model that is pre-trained on a more than $3000$M words corpora. It is popularly used as a feature extractor in NLP tasks.
By adding values as nodes and attributes as edges, which connect values and the entity, into the graph, we then can apply attention from the entity to attributes and use the attention score to compute the weighted average of attributes and values.
Following the Graph Attention Networks~\cite{velivckovic2018graph}, we define $\mathbf{h}^1_e$ as follows:
\begin{equation}
    \begin{aligned}
        \mathbf{h}^1_e & =\sigma(\sum_{j=1}^{n} \alpha_j \mathbf{W}_1[\mathbf{a}_j;\mathbf{v}_j]),          \\
        \alpha_j       & = \mathrm{softmax}(o_j)=\frac{\mathrm{exp}(o_j)}{\sum_{k=1}^n{\mathrm{exp}(o_k)}}, \\
        o_j            & =\mathrm{LeakyReLU}(\mathbf{u}^T[\mathbf{h}^0_e; \mathbf{a}_j]),
    \end{aligned}
\end{equation}
where $j\in \{1,\cdots, n\}$, $\mathbf{W}_1\in \mathbb{R}^{D_{h_1}\times (D_a + D_v)}$ and $\mathbf{u}\in \mathbb{R}^{(D_e + D_a)\times 1}$ are learnable matrices, $\sigma$ is the $\mathrm{ELU}(\cdot)$ function, and $\mathbf{h}^0_e$ is the initial entity feature.

\subsubsection{Mean Aggregator}
Mean aggregator layer utilizes the features of the target entity and its neighbors to generate the entity embedding. The neighbor entities of $e$ are defined by relation triples: $\mathcal{N}(e)=\{j\ \vert \ \forall (j, r, e)\in T^r\ or \ \forall (e, r, j)\in T^r, \forall r\in R\}$.
We aggregate the features of e's neighbor entities to gather alignment signal and learn the structural knowledge.
Given the hidden state $\mathbf{h}^{l-1}_{e}$ from the $l-1$ layer, the mean aggregator~\cite{hamilton2017inductive} is defined as:
\begin{equation}
    \small
    \mathbf{h}^l_e = \sigma(\mathbf{W}_l~\mathrm{MEAN}(\{\mathbf{h}^{l-1}_e\}\cup \{\mathbf{h}^{l-1}_j, \forall j\in \mathcal{N}(e)\}))
\end{equation}
where $\mathbf{W}_l\in \mathbb{R}^{D_{h_{l}}\times D_{h_{l-1}}}$ is a learnable matrix, $\mathrm{MEAN}(\cdot)$ returns the mean vector of the inputs, and $\sigma$ is the nonlinear function chosen as $\mathrm{ReLU}(\cdot)$.

\subsection{Graph Alignment}
Graph Alignment unifies the two KGs’ representations of each channel into a unified vector space by reducing the distance between the seed equivalent entities. We separately train the four channels and ensemble their outputs afterward for final evaluation (see Section~\ref{sec:ensemble}). 
Following~\citet{li2019semi}, we generate negative samples of $(e, e')\in \psi^{s}$ by searching the nearest entities of $e$ (or $e'$) in the entity embedding space. We denote the final output $\mathbf{h}^L_e$ of the channel $GC^k$ as the entity embedding $\textbf{e}^k$. For each channel $GC^k$, we optimize the following objective function:
\begin{equation}
    \small
    \begin{aligned}
        \mathcal{L}_{k} = \sum_{(e, e')\in \psi^{s}} & (\sum_{e_- \in \textrm{NS}(e)} [d(\mathbf{e}^k, \mathbf{e'}^{k}) - d(\mathbf{e}_-^k, \mathbf{e'}^k) + \gamma]_+ \\ + &\sum_{e'_-\in \mathrm{NS}(e')}[d(\mathbf{e}^k, \mathbf{e'}^k) - d(\mathbf{e}, \mathbf{e'}_-^k) + \gamma]_+)
    \end{aligned}
\end{equation}
where $\psi^s$ is the seed set of equivalent entities, $\mathrm{NS}(e)$ denotes the negative samples of $e$; $[\cdot]_{+} = \mathrm{max}\{\cdot, 0\}$, $d(\cdot, \cdot) = 1- \mathrm{cos}(\cdot, \cdot)$ is the cosine distance, and $\gamma$ is a margin hyperparameter.

\subsection{Channel Ensemble}
\label{sec:ensemble}
We use the entity embedding of each channel to infer the similarity matrices $\mathbf{S}^{k} \in \mathbb{R}^{|E|\times |E'|}$ ($k\in \{1,2,3,4\}$), where $\mathbf{S}^{k}_{e, e'}=\mathrm{cos}(\mathbf{e}^k, \mathbf{e'}^k)$ is the cosine similarity score between $e\in E$ and $e'\in E'$. We present two methods to ensemble the four matrices into a single similarity matrix $\mathbf{S}^*$ for final evaluation.

\noindent \textbf{Average Pooling.} Empirically, we assume that each channel has equal importance. We let $\mathbf{S}^*=\frac{1}{4}\sum_{k=1}^4 \tilde{\mathbf{S}}^{k}$, where $\tilde{\mathbf{S}}^k$ is the standardized $\mathbf{S}^k$:
\begin{equation}
    \tilde{\mathbf{S}}^{k} = \frac{\mathbf{S}^{k} - \mathrm{mean}(\mathbf{S}^{k})}{\mathrm{std}(\mathbf{S}^{k})}
\end{equation}

\noindent \textbf{SVM.} We utilize LS-SVM~\cite{suykens1999least} to learn the weights for each channel: $\mathbf{S}^{*}=\sum_{k=1}^{4}w_k \mathbf{S}^{k}$, where $\mathbf{w}=[w_1,w_2,w_3,w_4]$ is trained as follow:
\begin{equation}
    \begin{split}
        \mathcal{L}_{\mathrm{svm}} = C & \sum_{l=1}^{m} [y_l \cdot \mathrm{max}(0, 1-\mathbf{w}^T\mathbf{x}_l) + (1-y_l)\cdot \\
        & \mathrm{max}(0,1+\mathbf{w}^T\mathbf{x}_l)] + \frac{1}{2} \mathbf{w}^T \mathbf{w}
    \end{split}
\end{equation}
where $\mathbf{x}_l=[\mathbf{S}_{e, e'}^{1}, \mathbf{S}_{e, e'}^{2}, \mathbf{S}_{e, e'}^{3}, \mathbf{S}_{e, e'}^{4}]$ is a vector of sampled similarity scores. If $(e,e')\in \phi_s$, label $y_l=1$ , otherwise $y_l=0$.

\section{Experiments}
In this section, we compare AttrGNN with 12 baselines on the regular setting and our designed hard setting of EA. We also present an ablation study and a case study to evaluate attributes' and values' effects for EA.

\subsection{Experimental Settings}

\begin{table}[!t]
    \small
    \centering
    \begin{tabular}{cccc}
        \toprule
        \textbf{Datasets} & \textbf{\#Relation} & \textbf{\#Digital} & \textbf{\#Literal} \\
        \midrule
        DBP\tiny{ZH}      & 70k                & 182k           & 286k          \\
        DBP\tiny{EN}      & 95k                & 205k           & 291k          \\
        \midrule
        DBP\tiny{JA}      & 77k                & 156k           & 224k          \\
        DBP\tiny{EN}      & 93k                & 173k           & 267k          \\
        \midrule
        DBP\tiny{FR}      & 106k                & 165k           & 312k          \\
        DBP\tiny{EN}      & 116k                & 234k           & 218k          \\
        \midrule
        DWY\tiny{WD}      & 449k                & 362k          & 628k          \\
        DWY\tiny{DB}      & 463k                & 219k          & 403k          \\
        \midrule
        DWY\tiny{YG}      & 503k                & 115k         & 713k          \\
        DWY\tiny{DB}      & 429k                & 253k          & 506k          \\
        \bottomrule
    \end{tabular}%
    \caption{Triple numbers of datasets.
     \#Relation indicates the number of relation triples. The numbers of attribute triples that have digital values and literal values are denoted by \#Digital and \#Literal.}
    \label{tab:datasets}%
\end{table}%

\begin{table}[!t]
    \small
    \centering
    \begin{tabular}{|r|c|c|c|c|}
        \hline
                                                    & \textbf{Attr} & \textbf{Value} & \textbf{Name} & \textbf{Iter} \\ \hline
        MTransE~\shortcite{chen2016multilingual}    &               &                &               &               \\ \hline
        JAPE~\shortcite{sun2017cross}               & \checkmark    &                &               &               \\ \hline
        IPTransE~\shortcite{zhu2017iterative}       &               &                &               & \checkmark    \\ \hline
        AlignE~\shortcite{sun2018bootstrapping}     &               &                &               &               \\ \hline
        BootEA~\shortcite{sun2018bootstrapping}     &               &                &               & \checkmark    \\ \hline
        KDCoE~\shortcite{chen2018co}                &               &                &               & \checkmark    \\ \hline
        GCN-Align~\shortcite{wang2018cross}         & \checkmark    &                &               &               \\ 
        \hline
        MuGNN~\shortcite{cao2019multi}              &               &                &               &               \\ \hline
        \hline
        AttrE~\shortcite{trisedya2019entity}        & \checkmark    & \checkmark     & \checkmark    &               \\ \hline
        MultiKE~\shortcite{Zhang2019MultiviewKG}    & \checkmark    & \checkmark     & \checkmark    &               \\ \hline
        GraphMatch~\shortcite{Xu2019CrosslingualKG} &               &                & \checkmark    &               \\ \hline
        RDGCN~\shortcite{wu2019relation}            &               &                & \checkmark    &               \\ \hline
        AttrGNN (Ours)                              & \checkmark    & \checkmark     & \checkmark    &               \\ \hline
    \end{tabular}
    \caption{Characteristics of entity alignment models. The top part lists 8 models without utilizing entity names, and the bottom part lists 5 models with entity names. Attr and Value indicate the attributes and values from attribute triples; Name indicates entity names; and Iter indicates whether the model iteratively enlarge training set of equivalent entities.}
    \label{tab:baseline}%
\end{table}

\noindent \textbf{Datasets.} We test models on both cross-lingual and monolingual datasets: DBP15k~\cite{sun2017cross} and DWY100k~\cite{sun2018bootstrapping}. DBP15k includes three cross-lingual datasets collected from DBpedia: Chinese and English (DBP{\scriptsize ZH-EN}), Japanese and English (DBP{\scriptsize JA-EN}), French and English (DBP{\scriptsize FR-EN}). DWY100k contains two monolingual datasets: DBpedia and Wikidata (DBP-WD), DBpedia and YAGO (DBP-YG). The original DBP15k does not have attribute triples. Therefore we retrieve attribute triples from the DBpedia dump (2016-10). We then randomly sample 30\% of gold entity alignments for training and use the rest for testing. For DWY100k, we use the released attribute triples and the train/valid/test split of~\citet{Zhang2019MultiviewKG}. We show the number of relation/attribute triples for each dataset in Table~\ref{tab:datasets}.

\begin{table*}[ht!]
    \small
    \setlength{\tabcolsep}{4.0pt}
    \renewcommand{\arraystretch}{1.0}
    \centering
    \begin{tabular}{c|ccc|ccc|ccc}
        \toprule
        \multirow{2}[4]{*}{\textbf{Methods}} & \multicolumn{3}{c}{\textbf{DBP{\tiny ZH-EN}}} & \multicolumn{3}{c}{\textbf{DBP{\tiny{JA-EN}}}} & \multicolumn{3}{c}{\textbf{DBP{\tiny FR-EN}}}                                                                                                       \\
        \cmidrule{2-10}   & \textbf{H@1}   & \textbf{H@10}  & \textbf{MRR}   & \textbf{H@1}   & \textbf{H@10}  & \textbf{MRR}   & \textbf{H@1}   & \textbf{H@10}  & \textbf{MRR}   \\
        \midrule
        MTransE           & 30.83          & 61.41          & 0.364          & 27.86          & 57.45          & 0.349          & 24.41          & 55.55          & 0.335          \\
        JAPE              & 41.18          & 74.46          & 0.490          & 36.25          & 68.50          & 0.476          & 32.39          & 66.68          & 0.430          \\
        AlignE            & 47.18          & 79.19          & 0.581          & 44.76          & 78.89          & 0.563          & 48.12          & 82.43          & 0.599          \\
        BootEA            & 62.94          & 84.75          & 0.703          & 62.23          & 85.39          & 0.701          & 65.30          & 87.44          & 0.731          \\
        GCN-Align         & 41.25          & 74.38          & 0.549          & 39.91          & 74.46          & 0.546          & 37.29          & 74.49          & 0.532          \\
        MuGNN             & 49.40          & 84.40          & 0.611          & 50.10          & 85.70          & 0.621          & 49.50          & 87.00          & 0.621          \\
        \midrule
        NameBERT          & 60.36          & 71.00          & 0.642          & 74.53          & 83.57          & 0.779          & 87.44          & 92.06          & 0.891          \\
        MultiKE$^*$       & 43.70          & 51.62          & 0.466          & 57.00          & 64.26          & 0.596          & 71.43          & 76.08          & 0.733          \\
        GraphMatch        & 67.93          & 78.48          & -              & 73.97          & 87.15          & -              & 89.38          & 95.24          & -              \\
        RDGCN         & 70.75          & 84.55          & 0.749*          & 76.74          & 89.54          & 0.812*          & 88.64          & 95.72          & 0.908*          \\
        \midrule
        AttrGNN\tiny{avg} & \textbf{79.60} & \textbf{92.93} & \textbf{0.845} & \textbf{78.33}          & \textbf{92.08}          & \textbf{0.834}          & 91.85          & 97.77          & 0.910          \\
        AttrGNN\tiny{svm} & 77.72          & 92.00          & 0.829          & 76.25 & 90.88 & 0.816 & \textbf{94.24} & \textbf{98.67} & \textbf{0.959} \\
        \bottomrule
    \end{tabular}%
    \caption{Overall performance on the regular setting of DBP15k. Models in the first part do not use \textit{Names} while models in the second part use \textit{Name}. * indicates results from our re-implementation using their source code.}
    \label{tab:cross}%
\end{table*}%

\noindent \textbf{Baselines.} We compare AttrGNN with 12 baselines. We summarize four common characteristics of EA models and mark the employed characteristic for each method in Table~\ref{tab:baseline}. Among them, AttrE and MultiKE use the same information as AttrGNN. We also construct a baseline NameBERT that only uses the BERT embedding of entity names to measure the similarity. For each model, we list the reported performance if available; otherwise, we run the source code to get the result.  
Following existing works~\cite{sun2018bootstrapping}, we employ Hits@N (\%, short as H@N) and Mean Reciprocal Rank (MRR) as the evaluation metrics. Higher Hits@N and MRR indicate better performance. 


\noindent \textbf{Training Details.}
We use BERT~\cite{devlin2019bert} to initialize the feature vector for each value. Specifically, given a value $v$ consisting of a sequence of tokens, we use the pre-trained \textit{bert-base-cased}\footnote{https://github.com/huggingface/transformers} to generate a sequence of hidden states and apply max-pooling to obtain a fixed length vector $\mathbf{v}$ as the initial value feature vector. We do not fine-tune the BERT so that the feature vectors can be cached for efficiency. Following~\citet{sun2017cross}, we use Google Translate to translate all values to English for cross-lingual datasets. We initialize the four channels defined in Section~\ref{sec:channel} as follows. 
For the {\it Name} channel, we initialize the entity features using the BERT embedding of entity names. For the {\it Literal}, {\it Digital}, and {\it Structure} channels, we use randomly initialized the $128$ dimensional vectors as the entity and attribute features. We use Adagrad~\cite{duchi2011adaptive} as the optimizer. 
For each entity, we choose maximum $20$ or $3$ attribute triples based on GPU memory. For Graph Alignment, we choose $25$ negative samples for each entity. We use $16$ negative samples for each positive sample in the SVM ensemble model. We grid search the best parameters for each GNN channel on the valid set (if available) in the following range: learning rate $\{0.001, 0.004, 0.007\}$, L2 regularization $\{10^{-4}, 10^{-3}, 0\}$. We set $\gamma = 1.0$. We train each channel for $100$ epochs. For the SVM in Channel Ensemble, we search for $C$ in range $\{10^{-6}, 10^{-5}, 10^{-4}, 10^{-3}, 10^{-2}, 10^{-1}\}$.  The experiments are conducted on a server with two 6-core 2.40ghz CPUs, one TITAN X, and 128 GB memory. On DBP15k, the \textit{Literal}/\textit{Digital}/\textit{Name} channel costs less than 20 minutes for a grid search, and \textit{Structure} channel costs less than 5 minutes. 

\subsection{Overall Performance}
We report the results in two settings: \textit{regular setting}, \textit{i.e.}, the setting used in the previous entity alignment works; and \textit{hard setting}, where we construct a harder test set for objective evaluation. 

\subsubsection{Regular Setting}

\noindent \textbf{Cross-lingual Dataset.} Table~\ref{tab:cross} shows the overall performance on DBP15K. We can see that: 

1. As compared to the second best model, AttrGNN achieves significant performance improvements of $5.10\%$ for Hits@1 and $0.056$ for MRR on average. This demonstrates the effectiveness of AttrGNN in integrating both attribute triples and relation triples. 

2. NameBERT, which only uses entity names, performs better than models without using names in most cases. This demonstrates our observations that (1) the datasets are name-biased; and (2) the evaluation result cannot reflect true EA performance in real-world situation.
Specifically, NameBERT performs better on DBP{\scriptsize FR-EN} than that on DBP{\scriptsize JA-EN} and DBP{\scriptsize ZH-EN}, which indicates a higher name-bias on DBP{\scriptsize FR-EN}. The reason is the better translation quality between French and English.

3. AttrGNN's performance improvement over baselines is higher on DBP{\scriptsize ZH-EN} ($8.85\%$) than those on DBP{\scriptsize JA-EN} ($1.59\%$) and DBP{\scriptsize FR-EN} ($4.86\%$). The primary reason is that on DBP{\scriptsize ZH-EN}, different channels of features complement each other better than those on DBP{\scriptsize JA-EN} and DBP{\scriptsize FR-EN}. 
The ratios\footnote{We test the ratios of two models’, \textit{i.e.}, the \textit{Name} channel and the ensemble of the other three channels, complementary correct predictions.} of complementary features on DBP{\scriptsize ZH-EN}/DBP{\scriptsize JA-EN}/DBP{\scriptsize FR-EN} are $19\%/9\%/5\%$. Thus, we benefit the most on DBP{\scriptsize ZH-EN} from the ensemble.

4. The SVM ensemble strategy performs better than average pooling on DBP{\scriptsize FR-EN}. On DBP{\scriptsize FR-EN}, the performances of AttrGNN channels are imbalanced: the \textit{Name} channel performs much better than other channels, as shown by the performance gap between NameBERT and baselines without names on these datasets. In these imbalanced cases, SVM performs better because it can adjust the weights of channels. However, we can not explain that the SVM strategy performs worse that average pooling on DBP{\scriptsize ZH-EN} and DBP{\scriptsize JA-EN}. In fact, the integration of the various KG features is an open problem. We leave that as a future work.

\noindent \textbf{Monolingual Dataset.} We evaluate models on this monolingual setting to inspect the name-bias level when there is no translation error. Table~\ref{tab:dwy100k} shows the performance on DWY100K. The overall performance is similar to that on DBP15k, on which AttrGNN achieves the best performance. There are three major observations:

1. NameBERT achieves nearly $100\%$ Hits@1 on DBP-YG, which shows more severe name-bias than that on the cross-lingual dataset. The reason is that both DBpedia and YAGO are derived from Wikipedia, resulting in that $77.60\%$ of the released equivalent entities have exactly the same names while the rest have very similar names, \textit{e.g.}, \textit{George B. Rodney} and \textit{George B Rodney}. This results dose not indicate that EA is solved because EA is still challenging when integrating KGs from different domains, where entity names can be very different.

2. AttrE and MultiKE, which use entity names, do not perform well because of their agnostic of attribute importance. The crucial alignment signal from \textit{Name} is thus averaged away by other attribute triples (in DBpedia, each entity has 7-8 attribute triples in average). 

3. MultiKE performs better than AttrE because it particularly sets a ``Name View'' to incorporate names. However, MultiKE performs worse than NameBERT on DBP-YG and DBP15k (Table~\ref{tab:cross}), indicating that its inefficient combination of ``Name View'' and other views harms the performance.

\begin{table}[!t]
    \centering
    \small
    \setlength{\tabcolsep}{2.0pt}
    \renewcommand{\arraystretch}{1.0}
    \begin{tabular}{c|ccc|ccc}
        \toprule
        \multirow{2}[4]{*}{\textbf{Methods}} & \multicolumn{3}{c}{\textbf{DBP-WD}} & \multicolumn{3}{c}{\textbf{DBP-YG}}                                                                            \\
        \cmidrule{2-7}    & \textbf{H@1}   & \textbf{H@10}  & \textbf{MRR}   & \textbf{H@1}   & \textbf{H@10}   & \textbf{MRR}   \\
        \midrule
        MTransE           & 28.12          & 51.95          & 0.363          & 25.15          & 49.29           & 0.334          \\
        JAPE              & 31.84          & 58.88          & 0.411          & 23.57          & 48.41           & 0.320          \\
        IPTransE          & 34.85          & 63.84          & 0.447          & 29.74          & 55.76           & 0.386          \\
        BootEA            & 74.79          & 89.84          & 0.801          & 76.10          & 89.44           & 0.808          \\
        KDCoE             & 57.19          & 69.53          & 0.618          & 42.71          & 48.30           & 0.446          \\
        GCN-Align         & 47.70          & 75.96          & 0.577          & 60.05          & 84.14           & 0.686          \\
        MuGNN             & 61.60          & 89.70          & 0.714          & 74.10          & 93.70           & 0.810          \\
        \midrule
        NameBERT          & 83.32          & 90.15          & 0.860          & 99.85          & 99.99           & 0.999          \\
        AttrE             & 38.96          & 66.77          & 0.487          & 23.24          & 42.70           & 0.300          \\
        MultiKE           & 91.86          & 96.26          & 0.935          & 88.03          & 95.32           & 0.906          \\
        \midrule
        AttrGNN\tiny{avg} & \textbf{96.08} & \textbf{98.86} & \textbf{0.972} & 99.89          & 99.99           & 0.999          \\
        AttrGNN\tiny{svm} & 85.50          & 93.73          & 0.884          & \textbf{99.96} & \textbf{100.00} & \textbf{1.000} \\
        \bottomrule
    \end{tabular}%
    \caption{Overall performance on DWY100K. The performance of AttrE is reported in ~\citet{Zhang2019MultiviewKG}.}
    \label{tab:dwy100k}%
\end{table}
\begin{table*}[htbp!]
    \centering
    \small
    \begin{tabular}{c|ccc|ccc|ccc}
        \toprule
        \multirow{2}[4]{*}{\textbf{Methods}} & \multicolumn{3}{c}{\textbf{DBP{\tiny ZH-EN}}} & \multicolumn{3}{c}{\textbf{DBP{\tiny JA-EN}}} & \multicolumn{3}{c}{\textbf{DBP{\tiny FR-EN}}}                                                                                                       \\
        \cmidrule{2-10}   & \textbf{H@1}   & \textbf{H@10}  & \textbf{MRR}   & \textbf{H@1}   & \textbf{H@10}  & \textbf{MRR}   & \textbf{H@1}   & \textbf{H@10}  & \textbf{MRR}   \\
        \midrule
        JAPE              & 34.97          & 56.63          & 0.451          & 31.07          & 52.03          & 0.410          & 25.30          & 48.29          & 0.361          \\
        AlignE            & 40.09          & 69.94          & 0.501          & 37.42          & 69.19          & 0.479          & 38.01          & 71.28          & 0.492          \\
        BootEA            & 51.26          & 74.60          & 0.593          & 49.31          & 74.64          & 0.578          & 51.28          & 76.93          & 0.603          \\
        GCN-Align         & 36.59          & 64.66          & 0.464          & 33.94          & 65.30          & 0.448          & 30.32          & 63.69          & 0.414          \\
        MuGNN             & 40.64          & 74.58          & 0.521          & 39.86          & 75.33          & 0.515          & 40.71          & 78.26          & 0.531          \\
        \midrule
        NameBERT          & 38.36          & 55.06          & 0.444          & 60.03          & 74.47          & 0.654          & 79.02          & 86.89          & 0.820          \\
        MultiKE           & 27.92          & 35.21          & 0.306          & 48.18          & 55.68          & 0.509          & 64.69          & 69.54          & 0.665          \\
        GraphMatch        & 50.06          & 66.93          & -              & 60.26          & 71.78          & -              & 83.50          & 90.47          & -              \\
        RDGCN             & 60.44          & 76.60          & 0.662          & 68.19          & 83.77          & 0.737          & 82.87          & 93.12          & 0.866          \\
        \midrule
        AttrGNN\tiny{avg} & \textbf{66.21} & \textbf{81.81} & \textbf{0.719} & 75.72          & 88.76          & 0.805          & 86.41          & 94.67          & 0.894          \\
        AttrGNN\tiny{svm} & 65.90          & 81.16          & 0.716          & \textbf{77.39} & \textbf{90.33} & \textbf{0.821} & \textbf{88.64} & \textbf{95.64} & \textbf{0.912} \\
        \bottomrule
    \end{tabular}%
    \caption{Overall performance on the hard setting of DBP15k.}
    \label{tab:dbp15k_new}
\end{table*}
\subsubsection{Hard Setting}
In the hard setting, we aim to carry out a more objective evaluation of EA models on a harder test set. We first introduce how to construct the test set and then present the results and discussion.

\noindent \textbf{Build Harder Test Set.} Let $E_s$ and $E'_s$ be the set of known aligned entities in $G$ and $G'$. First, we compute the similarity matrix $\mathbf{S}$ via NameBERT; each element $\mathbf{S}_{e, e'}$ denotes the similarity between the entity pair $e\in E_s$ and $e'\in E'_s$. Second, we sort each row of $\mathbf{S}$ in descending order, by ranking $(e,e')$ higher when there is less similarity in their names.
Finally, we pick the highest-ranked $60\%$ of equivalent entity pairs as the test set. The train set ($30\%$) and the valid set ($10\%$) are then randomly selected from the remaining set of data. We construct harder test set for the cross-lingual dataset only, because it is impractical to find equivalent entity pairs whose entities have very different names on the monolingual dataset, as shown by the performance of NameBERT in Table~\ref{tab:dwy100k}. 

\begin{figure}[!t]
    \centering
    \includegraphics[width=0.95\columnwidth]{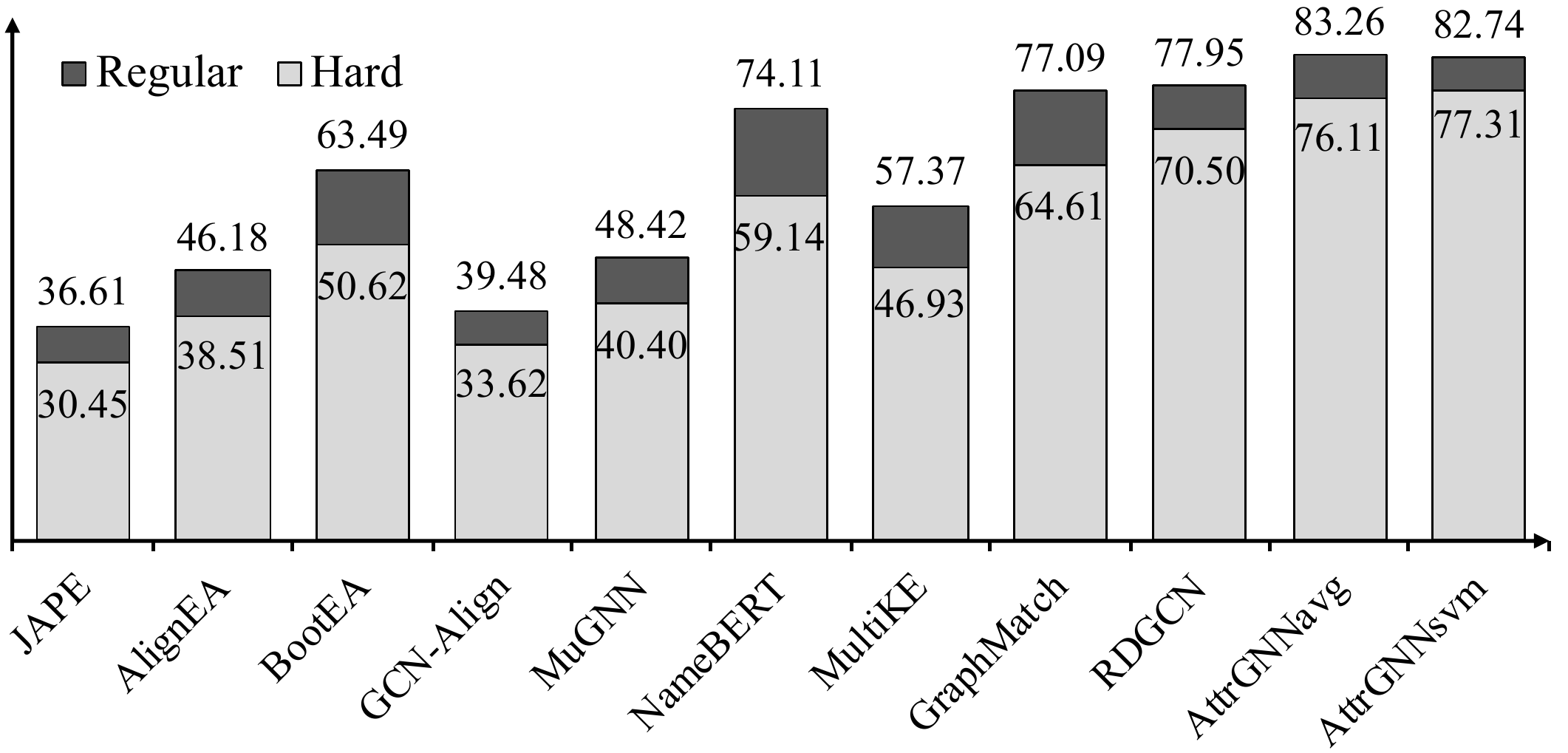}
    \caption{Average Hits@1 (\%) of models under the regular setting and the hard setting on DBP15k.}
    \label{fig:drop}
\end{figure}

\noindent \textbf{Discussion.} We implement AttrGNN and eight best-performed baselines with their source codes on the hard setting. Table~\ref{tab:dbp15k_new} shows the overall performance. We observe general performance drop in Hit@1 on DBP15k for all models, as shown in Figure~\ref{fig:drop}. There are three major observations: 

1. AttrGNN still achieves the best performance, demonstrating the effectiveness of our model. However, the performance of AttrGNN has degraded by around $6\%$ for Hits@1. This degradation indicates that the practical application of EA is still challenging and worth exploration.

2. AttrGNN shows the lowest degradation in performance among all the models with entity names. This stable performance demonstrates that incorporating attributes and values is important when the dataset is no longer biased to name.


3. Except for the iterative model, {\it i.e.}, BootEA, the performance of models without using entity names exhibits less performance drop than the models with names. The iterative model's performance degrades more because the harder dataset weakens the snowball effect
\footnote{https://en.wikipedia.org/wiki/Snowball\_effect}
when iteratively enlarging the seed set of equivalent entities.

\subsection{Ablation Study}

\begin{figure}[htb]
    \centering
    \includegraphics[width=0.95\columnwidth]{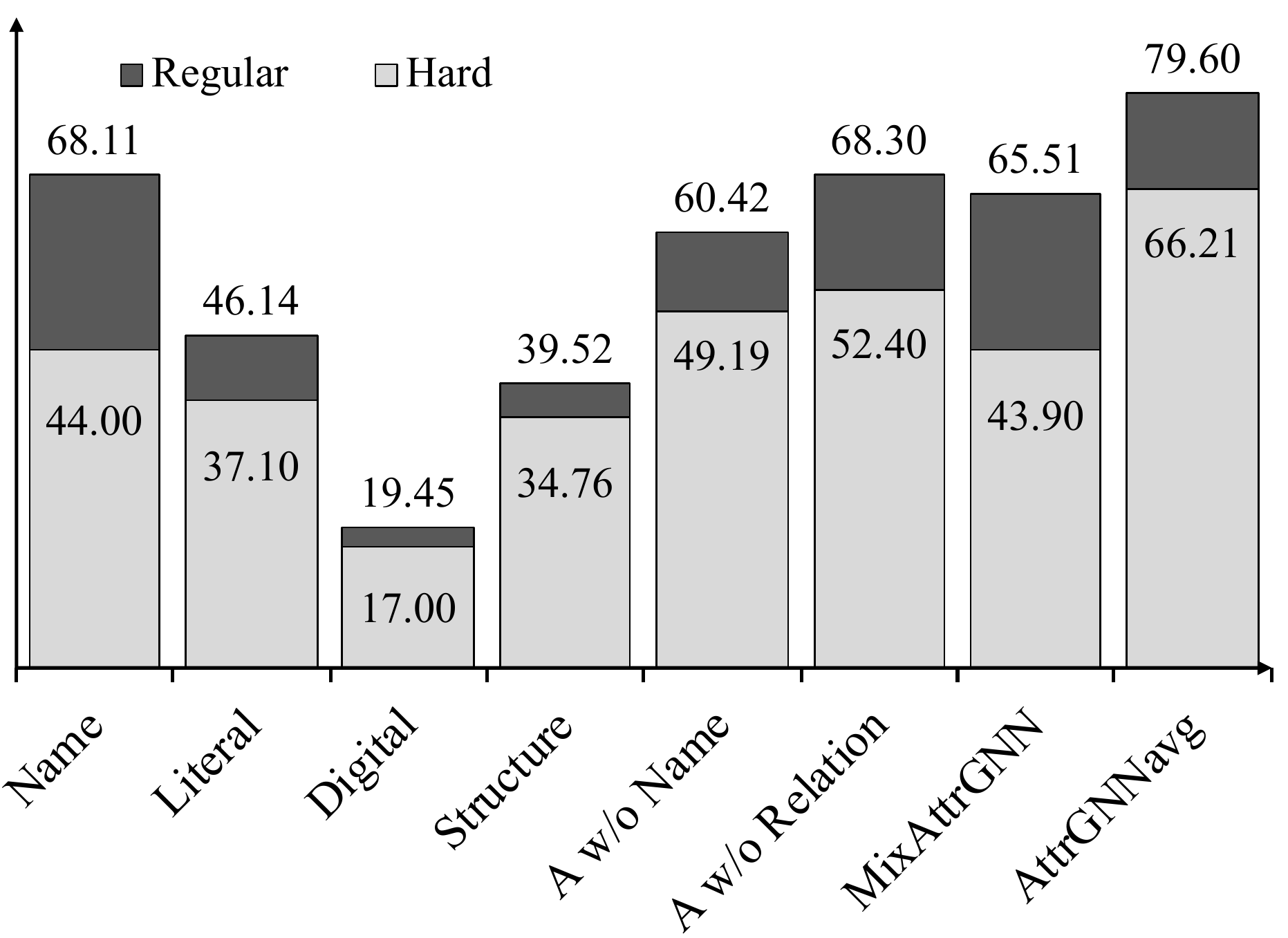}
    \caption{Ablation study on DBP{\scriptsize ZH-EN} (Hits@1 \%).}
    \label{fig:abl}
\end{figure}

We conduct an ablation study on the performance of each AttrGNN channel, AttrGNN{\scriptsize avg} without using the \textit{Name} channel (A w/o Name), AttrGNN without using relation triples (A w/o Relation), and AttrGNN without graph partition (MixAttrGNN) (Figure~\ref{fig:abl}). A w/o Relation is to ensemble NameBERT and one-layer Literal and Digital channels. There are three major observations: 

1. The {\it Literal} and {\it Structure} channels' performances are close to the {\it Name} channel under the hard setting. This demonstrates the importance to explore non-name features, including other attributes and relation, for practical EA.

2. Compared to MixAttrGNN, our simple graph partition strategy achieves promising improvement. The reason is that graph partition enables model to measure the similarity of different attributes differently.

3. The \textit{Digital} channel's performance is poor because it is challenging to learn the numerical calculation with the supervision of entity alignment. We thus leave it as future work.

4. Our full model significantly outperforms the Structure channel and the A w/o relation, which are the models with only relation/attribute features. This demonstrates the necessity of considering both relation and attribute triples for EA.

\begin{table}[htb]
    \centering
    \small
    \begin{tabular}{ccc}
        \toprule
        Score & Attribute          & Value               \\
        \midrule
        \multicolumn{3}{l}{\textbf{English Entity: Georgia (U.S. state)}} \\
        .109  & postalabbreviation & GA                  \\
        .039  & former             & Province of Georgia \\
        .037  & flag               & Flag of Georgia.svg \\
        .028  & arearank           & 24                  \\
        ...   &                    &                     \\
        .020  & senators           & David Perdue        \\
        .020  & governor           & Nathan Deal         \\
        .019  & motto              & Wisdom, Justice...  \\
        \midrule
        \multicolumn{3}{l}{\textbf{Chinese Entity: Georgia}}              \\
        .144  & postalabbreviation & GA                  \\
        .048  & flag               & Flag of Georgia.svg \\
        .041  & fullZhName         & Georgia             \\
        .037  & arearank           & 24                  \\
        ...   &                    &                     \\
        .026  & officiallang       & English             \\
        .026  & admittancedate     & 1788                \\
        .025  & totalarea          & 154077              \\        
        \bottomrule
    \end{tabular}%
    \caption{Attributes and values for the entity ``Georgia (U.S. state)'' from the English and Chinese DBpedia. Attributes are sorted in descending order according to the attention score. Chinese texts are translated.}
    \label{tab:case}
\end{table}%

\subsection{Case Study of Attributes and Values}
We give a qualitative analysis of how attribute triples contribute to EA in this case study. Table~\ref{tab:case} shows an equivalent entity pair that NameBERT fails to align, but AttrGNN aligns it by taking alignment signal from attributes and values. We observe that most of the top-ranked attributes have similar values between two KGs. In this case, the similar values include three literal strings, \textit{e.g.}, \textit{GA}, \textit{Flag of Georgia} and \textit{Seal of Georgia}, and a number, \textit{e.g.} \textit{24}. Meanwhile, the values that are not shared in both KGs are assigned low attention weights and filtered out. As similar cases are commonly observed, we conclude that -- attributes determine the importance of values, and values provide discriminative signals. In other words, the attributes whose values are unique are ranked higher, \textit{e.g.}, \textit{postalabbreviation} that denotes the unique postal abbreviation of provinces. The value of the lowest-ranked attributes may have different forms in different KGs. For example, the attention weight of \textit{totalarea} is small, because English KG and Chinese KG use different units of area (square mile in English DBpedia and square kilometer in Chinese DBpedia). 


\section{Conclusion and Future Work}
We propose a novel EA model (AttrGNN) and contribute a hard experimental setting for practical evaluation. AttrGNN can integrate both attribute and relation triples with varying importance for better performance. Experimental results under the regular and hard settings present significant improvements of our proposed model, and the severe dataset bias can be effectively alleviated in our proposed hard setting.

In the future, we are interested in replacing BERT with knowledge enhanced and number sensitive text representations models~\cite{cao2017bridge, geva2020injecting}.

\section*{Acknowledgments}
This research is supported by the National Research Foundation, Singapore under its International Research Centres in Singapore Funding Initiative. Any opinions, findings and conclusions or recommendations expressed in this material are those of the author(s) and do not reflect the views of National Research Foundation, Singapore.




\bibliography{emnlp2020}
\bibliographystyle{emnlp2020}
\end{document}